\documentclass[3p,twocolumn]{elsarticle}

\usepackage{lineno}
\usepackage{amsmath,graphicx,stmaryrd}
\usepackage{array}
\usepackage{slashbox}
\usepackage{fixltx2e}
\usepackage[caption=false]{subfig}
\modulolinenumbers[5]

\journal{Speech Communication}

\begin{document}

\begin{frontmatter}

\title{Semi-supervised acoustic model training for speech with code-switching}
\tnotetext[mytitlenote]{This work was performed while the first author was on a research visit to SRI International.}

\author[run,sri]{Emre Y\i lmaz\corref{cor1}}
\ead{e.yilmaz@let.ru.nl}
\author[sri]{Mitchell McLaren}
\ead{mitchell.mclaren@sri.com}
\author[run]{Henk~van~den~Heuvel}
\ead{h.vandenheuvel@let.ru.nl}
\author[run]{David~A.~van~Leeuwen}
\ead{d.vanleeuwen@let.ru.nl}
\cortext[cor1]{Corresponding author}
\address[run]{CLS/CLST, Radboud University, Erasmusplein 1, 6525 HT Nijmegen, Netherlands}
\address[sri]{Speech Technology and Research Laboratory, SRI International, 333 Ravenswood Avenue, Menlo Park, CA, USA}

\begin{abstract}
In the FAME! project, we aim to develop an automatic speech recognition (ASR) system for Frisian-Dutch code-switching (CS) speech extracted from the archives of a local broadcaster with the ultimate goal of building a spoken document retrieval system. Unlike Dutch, Frisian is a low-resourced language with a very limited amount of manually annotated speech data. In this paper, we describe several automatic annotation approaches to enable using of a large amount of raw bilingual broadcast data for acoustic model training in a semi-supervised setting. Previously, it has been shown that the best-performing ASR system is obtained by two-stage multilingual deep neural network (DNN) training using 11 hours of manually annotated CS speech (reference) data together with speech data from other high-resourced languages. We compare the quality of transcriptions provided by this bilingual ASR system with several other approaches that use a language recognition system for assigning language labels to raw speech segments at the front-end and using monolingual ASR resources for transcription. We further investigate automatic annotation of the speakers appearing in the raw broadcast data by first labeling with (pseudo) speaker tags using a speaker diarization system and then linking to the known speakers appearing in the reference data using a speaker recognition system. These speaker labels are essential for speaker-adaptive training in the proposed setting. We train acoustic models using the manually and automatically annotated data and run recognition experiments on the development and test data of the FAME! speech corpus to quantify the quality of the automatic annotations. The ASR and CS detection results demonstrate the potential of using automatic language and speaker tagging in semi-supervised bilingual acoustic model training.
\end{abstract}

\begin{keyword}
Automatic speech recognition \sep code-switching \sep speaker and language recognition \sep bilingual acoustic modeling \sep semi-supervised training \sep Frisian language
\end{keyword}

\end{frontmatter}

%\linenumbers

\section{Introduction}
\label{sec:intro}

In linguistics, language interaction among minority languages spoken in multilingual societies has been well studied from the linguistic perspective~\cite{weinreich1953,myers1989,auer1998,muysken2000,thomason2001,bullock2009}. This interaction takes various forms including phonological, morphological, syntactic and lexical changes consequent to various linguistic phenomena such as word borrowing, interference and relexification. One prominent mechanism induced in the interacting languages is code-switching (CS), which is defined as the continuous alternation between two languages in a single conversation. These spontaneous language switches in a single conversation are prominent in multilingual societies in which minority languages are influenced by the majority language or majority languages are influenced by \textit{lingua francas}, such as English and French.

West Frisian (Frisian henceforth) is a language spoken language spoken in the northern provinces of the Netherlands with approximately half a million bilingual speakers, and can be considered under-resourced from a speech technology perspective. These speakers switch between the Frisian and Dutch languages in daily conversations. In the FAME! Project, these language switches are studied, and their influence on modern automatic speech recognition (ASR) systems is explored with the objective of building a robust recognizer that can handle this phenomenon. The main focus has been developing robust acoustic models capable of operating on the bilingual speech with a reasonable ASR and CS detection accuracy with the ultimate goal of performing spoken document retrieval from broadcast archives containing Frisian-Dutch speech, covering a period of over 50 years~\cite{yilmaz2016_4}. Retrieving documents from these archives with decent accuracy is not only convenient for journalists, but it also reveals clues about various linguistic phenomena including language variation and CS trends in Frisian over years~\cite{dijkstra2017}. This is of particular interest to linguists as it enables access to a large amount of longitudinal bilingual speech and video data.

The impact of CS and other kinds of language switches on ASR systems~\cite{cetinoglu2016} has recently received research interest, resulting in several robust acoustic modeling~\cite{stemmer2001,lyu2006,vu2012,modipa2013,lyudovyk2014,wu2014,yilmaz2016_2,amazouz2017} and language modeling~\cite{li2012,adel2013,adel2014,westhuizen2017} approaches for CS speech. Given that CS involves more than one language, automatic language recognition (LR) could foreseeably assist in ASR of CS speech~\cite{weiner2012,lyu2013,yeong2014,mabokela2014}. One fundamental approach is assigning language labels in advance with the help of an LR system and then performing recognition of each language separately using a monolingual ASR system at the back-end. Such systems may suffer from error propagation from the language recognition front-end to the ASR back-end, especially for low-resourced languages and similar language pairs. To alleviate the adverse effect of irreversible language tag errors on ASR accuracy, all-in-one ASR approaches for CS speech have also been proposed (e.g., \cite{lyu2006,lyudovyk2014,yilmaz2016_2}), that do not directly incorporate an LR system.

Semi-supervised acoustic model training techniques, in which a small amount of reference data with annotations is used to bootstrap an ASR system that provides training labels for untranscribed speech data, have been researched intensively and various training strategies and data selection criteria have been proposed~\cite{gauvain1998,lamel2002,riccardi2005,yu2010,tsutaoka2012,liao2013,li2016,lileikyte2016,quitry2016}. These techniques are especially useful for increasing the available training data while building ASR systems for under-resourced languages with minimal amount of manually annotated data. Broadcast archives have been a popular source for extracting untranscribed speech data and this data is automatically transcribed using a basic ASR system trained on the already existing resources. The automatically transcribed data is later used for improving the initial acoustic models. The previous work in this line of research has focused on monolingual scenarios incorporating acoustic models operating on a single language. 

In this paper, we extend this idea to bilingual scenarios where the acoustic model is expected to recognize speech containing CS. The main contribution of this paper is the application of language and speaker recognition systems to untranscribed bilingual speech data, in order to obtain improved automatic annotation in a semi-supervised bilingual acoustic model training scenario. For this purpose, we first apply a robust speech activity detector to the raw broadcast data to extract segments with speech only. These raw speech segments have been automatically annotated using language and speaker information with the help of a language and a speaker recognition system, respectively. 

Using language recognition during the automatic annotation process enables the use of monolingual ASR systems which perform better (especially for the high-resourced Dutch language) than the baseline bilingual ASR system trained multilingually using only the limited amount of reference data~\cite{yilmaz2016_4}. Moreover, the impact of the LR errors on the final ASR performance will be limited, since they do not directly induce ASR errors unlike the aforementioned use of a LR system at the front-end of an ASR setup. Finally, lattice rescoring applied to improve the automatic transcription quality is more effective using monolingual LMs as shown in~\cite{yilmaz2017_2} compared to using a bilingual LM as done in~\cite{yilmaz2017_1}.  

For the annotation of speaker identities in the raw broadcast data, one viable approach is to link the speakers~\cite{vanleeuwen2010,ferras2012} appearing in the reference data with known identities to the (pseudo) speaker labels provided by a speaker diarization system. Speaker linking is achieved by clustering all speakers appearing in the reference and raw broadcast data based on the speaker similarity scores assigned by a speaker recognition (SR) system~\cite{ghaemmaghami2016}. This is not only crucial for tagging the speaker information when presenting a retrieved spoken document, but also for learning speaker-dependent acoustic features and models for an improved automatic transcription quality. In~\cite{cerva2013}, significant improvements have been reported in the transcription quality of a large spoken archive by applying speaker-adaptive ASR using the speaker labels given by a speaker diarization system. In this work, we further investigate the impact of speaker linking followed by the unsupervised diarization stage.

Finally, we compare the ASR performance of different acoustic models obtained after single-stage and two-stage~\cite{thomas2012,swietojanski2012,heigold2013,huang2013,ghoshal2013,tuske2013,knill2013,
vu2014,das2015,mohan2015} training strategies using the manually and automatically annotated (combined) data. We have reported some improvements in ASR accuracy on Frisian-Dutch speech using the two-stage multilingual deep neural network (DNN) training scheme~\cite{yilmaz2016_4}. With the expectation of getting larger improvements in the ASR performance, we use the combined data in the second stage of the training for tuning the acoustic models to the target CS speech. 

\begin{figure*}
  \includegraphics[trim=0cm 0cm 0cm 4cm, width=6.4in]{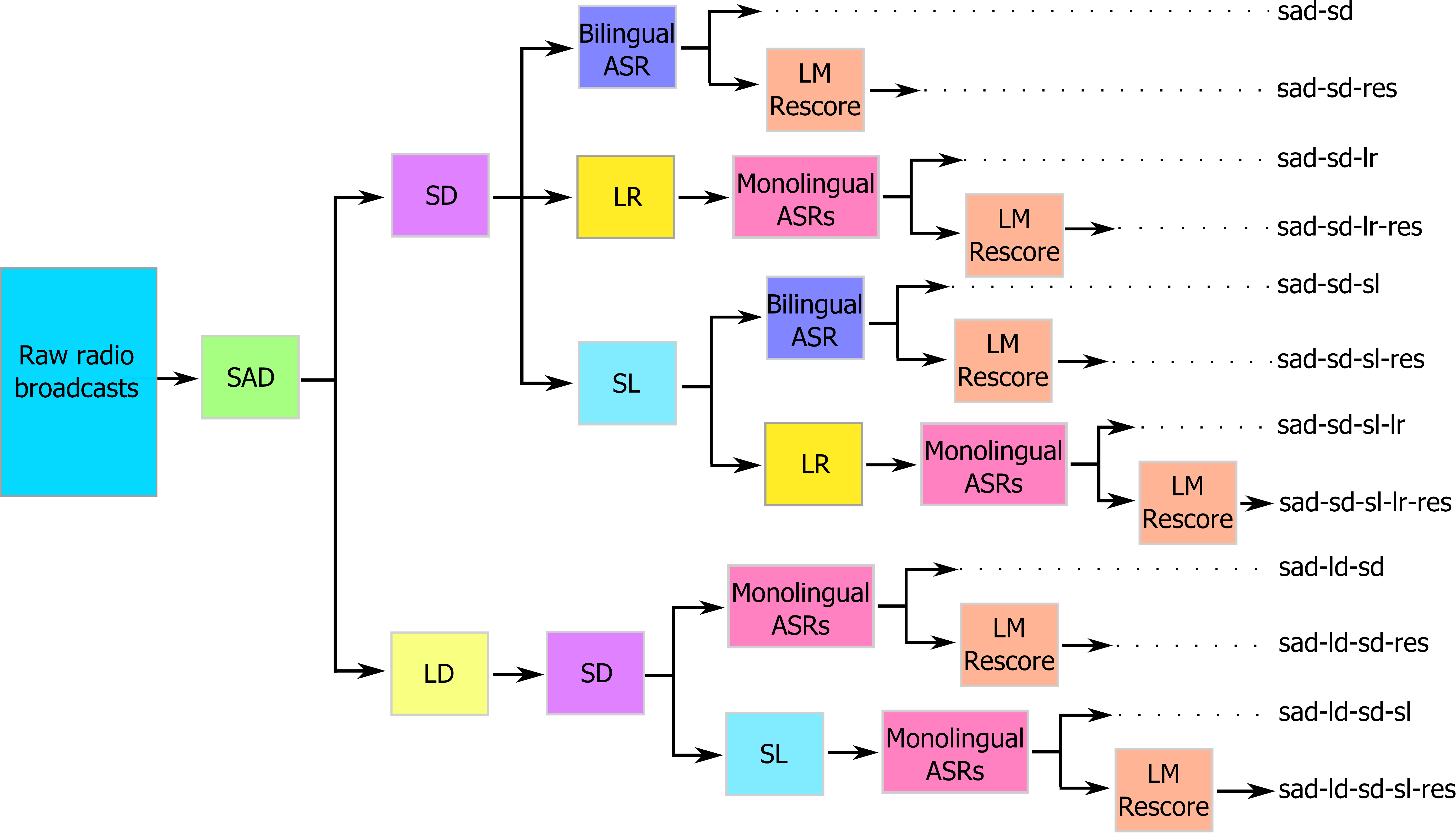}
  \caption{Overview of the automatic transcription systems (SAD: Speech activity detection; SD: Speaker diarization; LD: Language diarization; LR: Language recognition; SL: Speaker linking; LM: Language model)}
\label{fig:figure1}
\end{figure*}

This paper is organized as follows. Section~\ref{sec:database} introduces the demographics and the linguistic properties of the Frisian language and summarizes the Frisian-Dutch radio broadcast corpus collected for CS and longitudinal speech research. Section~\ref{sec:autotrans} details the proposed semi-supervised bilingual acoustic model training approaches using language and speaker recognition systems. The experimental setup is described in Section~\ref{sec:exps}, and the recognition results are presented in Section~\ref{sec:res}. A general discussion is given in Section~\ref{sec:dis} before concluding the paper in Section~\ref{sec:conc}.

\section{The Frisian Language and Frisian-Dutch Radio Broadcast Corpus}
\label{sec:database}

\subsection{The Frisian Language}
Frisian belongs to the North Sea Germanic language group, which is a subdivision of the West Germanic languages. Linguistically, three Frisian languages exist: (1) West Frisian, spoken in the province of Frysl\^{a}n in the Netherlands; (2) East Frisian, spoken in Saterland in Lower Saxony in Germany; and (3) North Frisian, spoken in the northwest of Germany, near the Danish border. Historically, Frisian shows many parallels with Old English~\cite{bremmer2009}. However, today the Frisian language is under the growing influence of the Dutch language due to long-lasting and intense language contact. Frisian has about half a million speakers. A recent study shows that about 55\% of all inhabitants of Frysl\^{a}n speak Frisian as their first language, which is about 330\,000 people~\cite{provinciefryslan}. All speakers of Frisian are at least bilingual, since Dutch is the main language used in education in Frysl\^{a}n.

\subsection{The FAME! Speech Corpus}

As was often emphasized during the recent research meetings on CS text/speech technology (the workshop in EMNLP 2016 and the special session in Interspeech 2017), this line of research suffers from the limited availability of (particularly spoken) data resources. Our contribution to expanding the limited amount of CS speech resources (e.g., \cite{chan2005,shen2011,li2012,imseng2012,dey2014,lyu2015,cetinoglu2017}) is the bilingual FAME! speech corpus~\cite{yilmaz2016}. This data contains radio broadcasts spoken in Frisian-Dutch extracted from the bilingual archive of the regional public broadcaster Omrop Frysl\^{a}n (Frisian Broadcast Organization). We first created a publicly available package of Frisian ASR resources including manually annotated speech data, a pronunciation lexicon, and a language model~\cite{yilmaz2016_3}. Subsequently, we created a speaker clustering and verification corpus, which added a large amount of raw broadcast data to the same manually annotated speech data~\cite{yilmaz2017_1}. 

The bilingual speech corpus designed for ASR experiments contains Frisian-only and Dutch-only utterances as well as mixed utterances with inter-sentential, intra-sentential and intra-word CS~\cite{myers1989}. To design an ASR system that can handle the language switches, a representative subset of recordings was extracted from this radio broadcast archive. These recordings include language switching cases and speaker diversity, and have a large time span (1966--2015). The content of the recordings is very diverse, including radio programs about culture, history, literature, sports, nature, agriculture, politics, society and languages.

The radio broadcast recordings were manually annotated and cross-checked by two bilingual native Frisian speakers. The annotation protocol includes three kinds of information: (1) the orthographic transcription containing the uttered words; (2) speaker details such as the gender, dialect, and name (if known); and (3) spoken language information. The language switches are marked with the label of the switched language. The total duration of the manually annotated radio broadcasts sums up to 18.5 hours with 14.5 hours of transcribed speech. The stereo audio data has a sampling frequency of 48 kHz and 16-bit resolution per sample. All data is subsampled to 16 kHz and reduced to single channel data. The available meta-information helped the annotators to identify the speakers and mark them either using speaker names or the same label (if the name is not known). Later, a manual check was performed by a colleague in the Fryske Akademy, who is also a bilingual native Frisian-Dutch speaker, to improve the quality of the speaker annotations with the help of Omrop Frysl\^{a}n employees. The FAME! speech corpus includes 334 identified and 120 unidentified speakers. In the corpus, 51 of the identified speaker appear at least in two different years. These speakers are mostly program presenters and celebrities appearing multiple times in different recordings over years. 10 speakers are labeled to speak in both languages.

Two kinds of language switches are observed in the broadcast data in the absence of segmentation information. First, a speaker may switch language in a single conversation (\textit{within-speaker switches}). Secondly, a speaker may be followed by another one speaking in the other language. For example, the presenter may narrate an interview in Frisian, while several excerpts of a Dutch-speaking interviewee are presented (\textit{between-speaker switches}). Both types of switches pose a challenge to ASR systems and must be handled carefully.

\section{Automatic Transcription of Raw Broadcast Data}
\label{sec:autotrans}
The automatic annotation schemes compared in this work are illustrated in Figure~\ref{fig:figure1}. In all approaches, speech-only segments are extracted from a large amount of raw broadcast data using a robust DNN-based speech activity detection (SAD) system~\cite{graciarena2016} to predict the posterior of the speech and non-speech classes at the DNN output. The speech-only segments belonging to the same recording are merged to extract a single speech segment for each raw recording. The following steps differ for each system and each component is detailed in the following sections.

\subsection{Front-End Processing}

\subsubsection{Speaker Diarization}
\label{sssec:sd}
The speech-only segments are labeled with speaker identities by using a speaker diarization (SD) system. The speaker diarization system used in this work attempts to cluster speakers in a recording such that each unique speaker is assigned to a single cluster and each cluster only has one speaker. The intent of diarization in the current context is to aid in speaker-adaptive training. Errors from diarization  (i.e, the allocation of speech to the wrong cluster) are expected to have limited impact on ASR since the errors will likely be due to similar sounding speakers.

The diarization technique used in this work is mostly based on the process defined in~\cite{sell2014}. Specifically, i-vectors~\cite{dehak2011} are first extracted from overlapping speech segments with a length of $N_{sd}$ seconds prior to an exhaustive comparison of segments with probabilistic linear discriminant analysis (PLDA)~\cite{prince2007} to provide a matrix of scores representing speaker similarity. These scores are then transformed into a distance matrix, with distances being computed as the opposite of the log-likelihood ratios (LLR) obtained acoustic model training with PLDA shifted by the maximum LLR obtained for any pair of samples. Consequently, the minimum distance is 0. Finally, hierarchical clustering with average linkage method is used to generate a clustering tree which is pruned to ensure that each cluster has a cophenetic distance no greater than a pre-defined threshold tuned on a held-out corpus of telephone conversations.

\subsubsection{Speaker Linking}
\label{sssec:sl}

Linking the speaker labels assigned by the SD system is a straightforward step towards improving the quality of the speaker labels assigned to the raw data. Improving the speaker label quality is crucial in this scenario, since improved speaker labels will yield more effective speaker-adaptive training of acoustic models. This is mainly due to the increased amount of data for the speakers appearing both in the reference and raw data (e.g., presenters and celebrities).

Motivated by these points, we also investigate the influence of speaker linking on the automatic annotation task. For this purpose, we use a speaker identification system trained on a large amount of multilingual data to assign speaker similarity scores. The performance of this system is found to be comparable with a similar system trained on the automatically labeled Frisian-Dutch data described in~\cite{yilmaz2017_3}. The speaker recognition system used in this paper is a single DNN-based system leveraging a hybrid alignment framework~\cite{mclaren2016} in which i-vectors are extracted for each audio segment and scored using standard PLDA to measure the speaker similarity given two speech segments. Further details about the speaker recognition system can be found in~\cite{mclaren2016}.

Similarity scores for all possible speaker pairs (i.e, the reference speakers with known speaker identities and the speakers with pseudo labels assigned by the SD system) are calculated. Based on these speaker similarity scores represented in the form of a distance matrix, speaker linking is performed by applying complete-linkage clustering as described in~\cite{ghaemmaghami2016}. A \textit{mild} level of linking is performed, as aggressive merging of the speaker labels conceivably has an inverse effect on the ASR performance, which has been also observed in the pilot experiments. All speaker labels associated to the same cluster are mapped to a new pseudo speaker label.

\subsubsection{Language Recognition}
\label{sssec:lr}

The speech segments are labeled with a language tag in two different stages to investigate the impact of different strategies on the automatic annotation quality. In the first case, language labeling is performed after assigning the speaker labels. In this case, segments belonging to the same speaker are merged with the assumption of monolingual speakers and the merged segments are labeled using a language recognition system. After assigning the labels, each utterance is automatically transcribed by the corresponding monolingual ASR system.

The language recognition system used in this work performs i-vector extraction at the front-end and a Gaussian backend scoring~\cite{song2013,ferrer2016}. The LR system is based on bottleneck features extracted from a DNN trained on a large amount English telephone data to predict tied tri-phone states (senones). These bottleneck features are used as input to the i-vector extraction~\cite{dehak2011}. The Dutch and Frisian language classes are modeled by Gaussians with shared covariance. The LLR of the detected languages is calculated using calibration with a multi-class objective function~\cite{brummer2006} trained on the same data used to estimate the Gaussian parameters.

\subsubsection{Language Diarization}
\label{sssec:ld}

Other language-specific strategies use a language diarization (LD) system before assigning speaker labels using the SD system. The speech-only segments provided by the SAD system are processed by the LD system, and language scores are assigned to the overlapping speech segments of $N$ seconds with a shift of $K$ seconds for $N$~\textgreater~$K$. The language scores are assigned using the LR system described in Section~\ref{sssec:lr}. For each segment of $K$ seconds, we apply majority voting among all language scores to decide on the assigned language label. The arguable choice of majority voting is motivated by the limited number of hypothesized language switches avoiding false alarms. Based on the output of the LD system, each recording is segmented at language switch instants, and each segment is recognized using the corresponding monolingual ASR system.

\subsection{Back-End Processing}

\subsubsection{Bilingual ASR}
\label{sssec:basr}

Baseline automatic transcription is achieved by using a multilingually trained system that is also used in~\cite{yilmaz2016_4}. Using language resources from high-resourced languages for the recognition of an under-resourced language is common practice~\cite{besacier2014,imseng2014,juan2015}. Multilingual data from closely related high-resourced languages (i.e., Dutch and English) is used for training DNN-based acoustic models to obtain more robust acoustic models against the language switches between the under-resourced Frisian language and Dutch. 

The multilingual DNN training scheme (henceforth referred to as ML DNN) resembles prior work~\cite{thomas2012,swietojanski2012,huang2013} and is achieved in two steps. First, the English and Dutch data are used together with the Frisian data in the first step to obtain more accurate shared hidden layers extracted from a DNN model with a separate softmax layer for English, Dutch and Frisian. After the first stage, the trilingual softmax layer used in the first step is replaced with one that is specific to the target recognition task, namely Dutch and Frisian. In the second step, the complete DNN is retrained bilingually to fine-tune the DNNs for the target CS Frisian and Dutch speech.

The choice of fully connected DNNs can be questioned rather than more sophisticated, e.g. convolutional or recurrent, neural network architectures for the evaluation of the proposed strategies. We have also investigated the performance of convolutional and recurrent models trained on both manually annotated and combined data in pilot experiments. Our choice of the fully connected DNN models is based on 1) their better performance than recurrent models (possibly due to lack of training data) and comparable performance to convolutional models on the development data, and 2) their lower computational requirements compared to the convolutional and recurrent models.

In this multilingual training scheme, the phones of each language are modeled separately (e.g., by appending a language identifier to every phone of a word based on the language of its lexicon). We refer the reader to~\cite{yilmaz2016_2} in which the impact of phone merging in the context of CS ASR is explored. The words in each lexicon are also tagged with a language identifier to enable evaluating CS detection accuracy.

The most likely hypothesis output by the recognizer is used as the reference transcription. After obtaining the transcriptions, the manually and automatically transcribed data are merged to obtain the combined Frisian-Dutch broadcast data, which is used for the training of the final bilingual acoustic models.
\begin{table*}
\centering
%\addtolength{\tabcolsep}{-1.95pt}
\caption{Data composition of different training setups used in the recognition experiments (in hours)}
\vspace{0.25cm}
\begin{tabular}{| l | c | c | c | c | c |} 
\hline
Training data & Annotation & Frisian & Dutch  & English & Total  \\
\hline\hline
FAME! train   & Manual &   8.5   &   3.0  &    -    & 11.5  \\
\hline
ML data~\cite{yilmaz2016_4}& Manual &   8.5   & 110.0  &  141.5  & 260.0\\
\hline\hline
sad-sd        & Automatic + Manual & \multicolumn{2}{c|}{137.0}& - & 137.0  \\
\hline
sad-sd-lr     & Automatic + Manual  &   77.0  &  60.0   &  -      &   137.0   \\
\hline
sad-ld-sd     & Automatic + Manual  &   81.5  &  49.5   &  -      &   131.0   \\
\hline
\end{tabular}
% \vspace{-0.3cm}
\label{tab:data}
\end{table*}
\subsubsection{Monolingual ASRs}
\label{sssec:mono}

The language-labeled segments are automatically transcribed using the monolingual resources at the back-end. Having one low-resourced and one high-resourced language as mixed languages, the monolingual ASR of the highly resourced language, Dutch in this case, has a higher ASR accuracy on the Dutch segments compared to a bilingual system, which is trained to recognize both languages. Moreover, the multilingual training approach described in Section~\ref{sssec:basr} can be applied for creating acoustic models that recognize Frisian speech only. This is achieved by using the Frisian speech only in the second stage of the training. Such a system also provides better recognition than a bilingual system and, hence, improved automatic transcriptions.

\subsubsection{Language Model Rescoring}
\label{sssec:lmres}

Lattice rescoring using a bilingual and two monolingual recurrent neural network (RNN) LMs has also been applied to extract alternative transcriptions. In~\cite{yilmaz2017_1}, the transcriptions extracted with and without the rescoring stage gave similar results. However, in that case, the rescoring was performed with a bilingual LM with a higher perplexity compared to the monolingual LMs used in this work. Therefore, we include the rescoring stage in the experiments, expecting to see improved transcription quality due to the lower perplexity of the monolingual LMs. 

\section{Experimental Setup}
\label{sec:exps}

\subsection{Data Corpora}

Details of all training setups are presented in Table~\ref{tab:data}. The training data of the FAME! speech corpus comprises 8.5 hours and 3 hours of speech from Frisian and Dutch speakers respectively. The development and test sets consist of 1 hour of speech from Frisian speakers and 20 minutes of speech from Dutch speakers each. All speech data has a sampling frequency of 16 kHz.

The raw radio broadcast data extracted from the same archive with the FAME! speech corpus consists of 256.8 hours of audio, including 159.5 hours of speech based on the SAD output. The speech-only segments are fed to either the speaker diarization system described in Section~\ref{sssec:sd} or the language diarization system described in Section~\ref{sssec:ld}. The amount of remaining data after the removal of very short segments for each approach is given in the last three rows of Table~\ref{tab:data}. It is important to note that the durations for each language given in the last two rows are based on the language tags assigned by the language recognition/diarization systems. 

The Dutch speech databases used for ML DNN training are the Dutch broadcast database~\cite{vanleeuwen2009_2} and the components of the CGN~\cite{cgn} corpus that have broadcast-related recordings. These databases contain  17.5 and 89.5 hours of Dutch data respectively. In addition to this, the English Broadcast News Database (HUB4) is used as the main source of English broadcast data. The amount of the English data extracted from both the 1996 and 1997 components of HUB4~\cite{hub4_1996,hub4_1997} is approximately 141.5 hours.

\subsection{Implementation Details}
\label{ssec:impdet}

\subsubsection{ASR details}
\label{sssec:asrdet}

The recognition experiments are performed using the Kaldi ASR toolkit~\cite{kaldi}. We train a conventional context dependent Gaussian mixture model-hidden Markov model (GMM-HMM) system with 40k Gaussians using 39 dimensional mel-frequency cepstral coefficient (MFCC) features including the deltas and delta-deltas to obtain the alignments for training a fully connected DNN-based acoustic model. A standard feature extraction scheme is used by applying Hamming windowing with a frame length of 25 ms and frame shift of 10 ms. DNNs with six hidden layers and 2048 sigmoid hidden units at each hidden layer are trained on the 40-dimensional feature-level maximum likelihood linear transformations (fMLLR)~\cite{gales1998} features with the deltas and delta-deltas. The DNN training is done using the same speaker-dependent features by mini-batch stochastic gradient descent with an initial learning rate of 0.008 and a minibatch size of 256. The time context size is 11 frames achieved by concatenating $\pm$5 frames. We further apply sequence training using a state-level minimum Bayes risk (sMBR) criterion~\cite{vesely2013}. 

The bilingual lexicon contains 110k Frisian and Dutch words. The number of entries in the lexicon is approximately 160k due to the words with multiple phonetic transcriptions. The phonetic transcriptions of the words that do not appear in the initial lexicons are learned by applying grapheme-to-phoneme (G2P) bootstrapping~\cite{davel2003,maskey2004}. The lexicon learning is done only for the words that appear in the training data using the G2P model learned on the corresponding language. We use the Phonetisaurus G2P system~\cite{novak2015} for creating phonetic transcriptions.

The details of the multilingually trained bilingual system that we used for the automatic transcription are found in~\cite{yilmaz2016_4}. The monolingual Frisian system used for automatically annotating the Frisian-labeled speech segments is also multilingually trained on the same data as the bilingual system, but retrained only on the Frisian training data. This system provides a WER of 32.7\% and 30.9\% on the Frisian component of the development and test sets respectively. The monolingual Dutch system is trained on 110 hours of Dutch speech from the CGN corpus~\cite{cgn}. Further details of the Dutch ASR system are available in~\cite{yilmaz2016_4}. This system provides a WER of 27.9\% and 23.8\% on the Dutch component of the development and test sets.

Standard bilingual and monolingual 3-gram with interpolated Kneser-Ney smoothing and RNN LMs are trained with 300 hidden units for recognition and lattice rescoring respectively. The 3-gram LMs are trained using the SRILM toolkit \cite{srilm}. The RNN LMs with gated recurrent units (GRU)~\cite{chung2014} and trained with noise contrastive estimation~\cite{chen2015} are obtained using the faster RNN LM training implementation\footnote{https://github.com/yandex/faster-rnnlm}. All language models are trained on a bilingual text corpus containing 37M Frisian and 8.8M Dutch words. The Frisian text is extracted from Frisian novels, news articles, Wikipedia articles and orthographic transcriptions of the FAME! training data. The bilingual 3-gram LM (RNN LM) has a perplexity of 260 (248) and 235 (215) on the transcriptions of FAME! development and test set respectively.

\subsubsection{SAD, SD, LR and SR Details}
\label{sssec:lsrdet}
SRI International's OLIVE software package~\cite{lawson2016} is used for speech activity detection, speaker diarization, speaker recognition and language recognition. The speech activity detection system uses 20-dimensional MFCC features with mean and variance normalization over the full waveform, and concatenated over a window of 31 frames. The resulting 620-dimensional feature vector is input to a DNN which consisted of two hidden layers of sizes 500 and 100. The output layer of the DNN consisted of two nodes trained to predict the posteriors for speech and nonspeech classes. These posteriors were converted into likelihood ratios using Bayes rule (assuming a prior of 0.5), and thresholded to make speech and non-speech decisions per frame.

The speaker diarization system uses segments of length $N_{sd}$ = 2 seconds with 50\% percent overlaps. For each recording, the diarization system assigns a maximum of eight different speaker labels after the speech activity detection and a maximum of four different speaker labels after language diarization.  

The speaker recognition system is based on the hybrid alignment framework presented in~\cite{mclaren2016}. It leverages bottleneck (BN) features extracted from a DNN to align traditional acoustic features in the Baum-Welch statistics calculation. The BN extractor is formed from 20-dimensional power normalized cepstral coefficients (PNCC)~\cite{kim2016} contextualized with principal component analysis-discrete cosine transform (pcaDCT)~\cite{mclaren2015} with a window of 15 frames to create 90 dimensional inputs to the DNN which are then mean and variance normalized using a sliding window of 3 seconds. The DNN is trained to discriminate 1933 senones using Fisher and Switchboard telephone data, and consists of 5 layers of 1200 nodes, except the fourth hidden layer which has 80 nodes and forms the bottleneck extraction layer. The first-order features, aligned with the bottleneck features and a 2048-component diagonal covariance UBM, are mel frequency cepstral coefficients (MFCC) of 20 dimensions also contextualized with pcaDCT using a 15 frame window with an output of 60 dimensions. In all cases, the PCA transform for pcaDCT are learned using a subset of the DNN training data. 400-dimensional i-vectors are extracted, which are length-normalized and reduced to a dimension of 200 using LDA before PLDA. This speaker recognition system had been tested beforehand on the FAME! speaker verification corpus \cite{yilmaz2017_3} to benchmark the speaker recognition performance on similar data extracted from the same radio archive. It had provided an equal error rate (EER) of 7.0\% on the 30-second test segments of the \textit{complete database} (please see \cite{yilmaz2017_3} for details) which is the most relevant scenario to the experiments presented in this work given that the system assigns similarity scores to relatively long-duration segments obtained by merging all segments labeled with the same pseudo-speaker.
 
\begin{table*}

\centering
\caption{Number of Frisian and Dutch words in the Frisian-only (fy), Dutch-only (nl) and code-switching (fy-nl) segments in the FAME! development and test sets}
\vspace{0.25cm}
%\addtolength{\tabcolsep}{-3.3pt}
\begin{tabular}{| c | c | c | c | c | c | c | c | c |}
\hline
 & \multicolumn{4}{c|}{Development} & \multicolumn{4}{c|}{Test} \\
\hline
 & fy & nl & fy-nl & all & fy & nl & fy-nl & all \\
\hline
 \# of Frisian words & 9190 & 0 & 2381 & 11,571 & 10,753 & 0 & 1798 & 12,551 \\
\hline
 \# of Dutch words & 0 & 4569 & 533 & 5102 & 0 & 3475 & 306 & 3781 \\
\hline
\end{tabular}
\label{tab:num}
\end{table*}
The speaker similarity scores are mapped to a distance matrix using an exponential function as described in~\cite{ghaemmaghami2016}. Speaker linking is achieved by using the hierarchical clustering implementation of the SciPy toolkit~\cite{scipy}. The total number of speakers in the reference data is 382. The speaker labels assigned by the SD system is 3734 for the sad-sd approach. After speaker linking (sad-sd-sl), these speakers are grouped into 1987 new speaker labels. For sad-ld-sd and sad-ld-sd-sl systems, these numbers are 3253 and 2550 respectively.

The language recognition system is based on bottleneck features extracted from a DNN trained on English telephone data from the Fisher and Switchboard datasets to predict more than 2000 tied tri-phone states (senones). Bottleneck features of 80 dimensions are used to extract 400 dimensional i-vectors~\cite{dehak2011} based on a 2048-component universal background model (UBM). The LR system is trained using segments of 30 seconds that are extracted from the training data of the FAME! corpus and it is evaluated on the development and test segments of 10 seconds to investigate LR accuracy before applying it to the raw data. This system has achieved an EER of 4.3\% on the described setting. The language priors are assumed to resemble the language distribution in the training data. Using more Dutch data from other databases for training resulted in lower LR accuracy on the evaluation data.

The language diarization is performed with a window length of $N=30\,$sec.\ and a shift $K=10\,$sec. As an initial step, we mainly aim for detecting \textit{between-speaker switches} (cf.~Section~\ref{sec:database}) in the raw broadcast data in this work. Therefore, a coarse LD with a shift on the order of 10 seconds is viable for this purpose. Investigating a finer LD that enables detecting intra-sentential and intra-word CS in raw data remains as future work.

\subsection{ASR and CS Detection Experiments}

The quality of different annotation strategies is quantified by evaluating the ASR and CS performance of the final bilingual acoustic models, trained using the corresponding annotations, on a separate set of manually annotated data used for testing purposes. There are two baseline systems which only uses the manually annotated training data: (1) the ASR system trained only on the manually annotated data (Man. Annot.), and (2) the multilingually trained DNN system \cite{yilmaz2016_4} which is retrained only on the manually annotated data (ML DNN (Man. Annot.)). Other ASR systems incorporate acoustic models trained on the combined (manually and automatically annotated) data and each approach is labeled with the system identities shown in Figure~\ref{fig:figure1}. 

\begin{figure*}[ht]
%\vspace{-0.35cm}
    \subfloat[Speaker Diarization without Language Recognition \label{subfig-1:sad_sd}]{%
    \includegraphics[trim=4.5cm 0.5cm 0cm 0cm, width=7in]{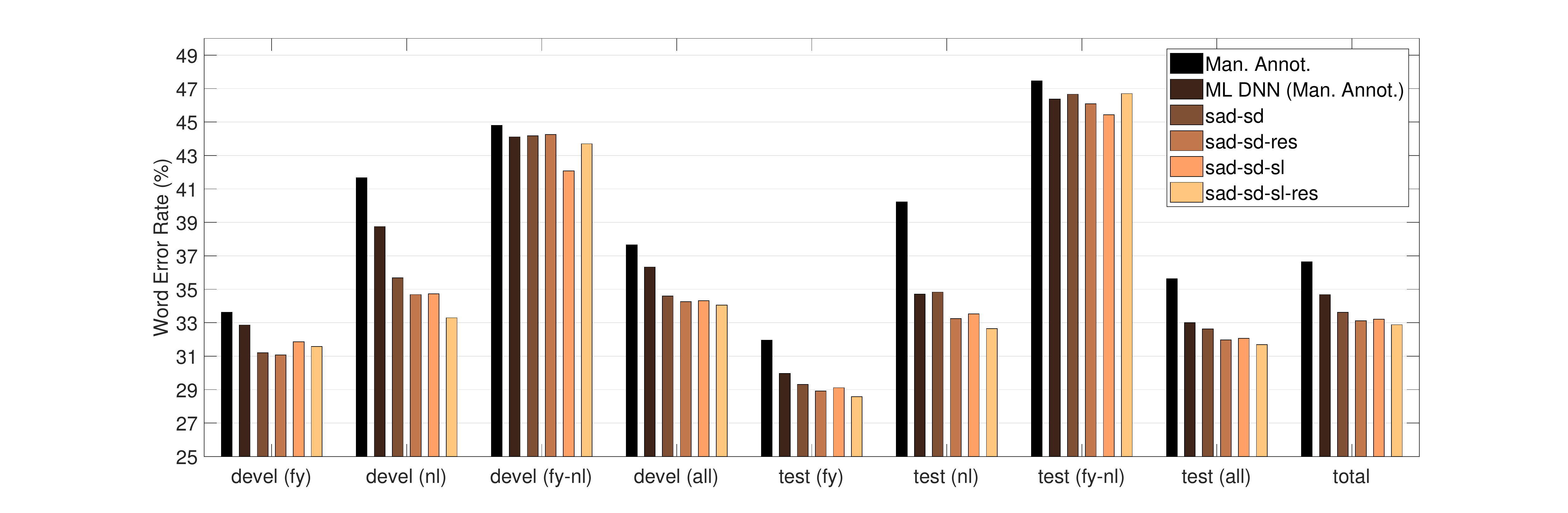}
    }
    \vspace{-0.35cm}
    \hfill
    \subfloat[Speaker Diarization with Language Recognition \label{subfig-2:sad_sd_lid}]{%
    \includegraphics[trim=4.5cm 0.5cm 0cm 0cm, width=7in]{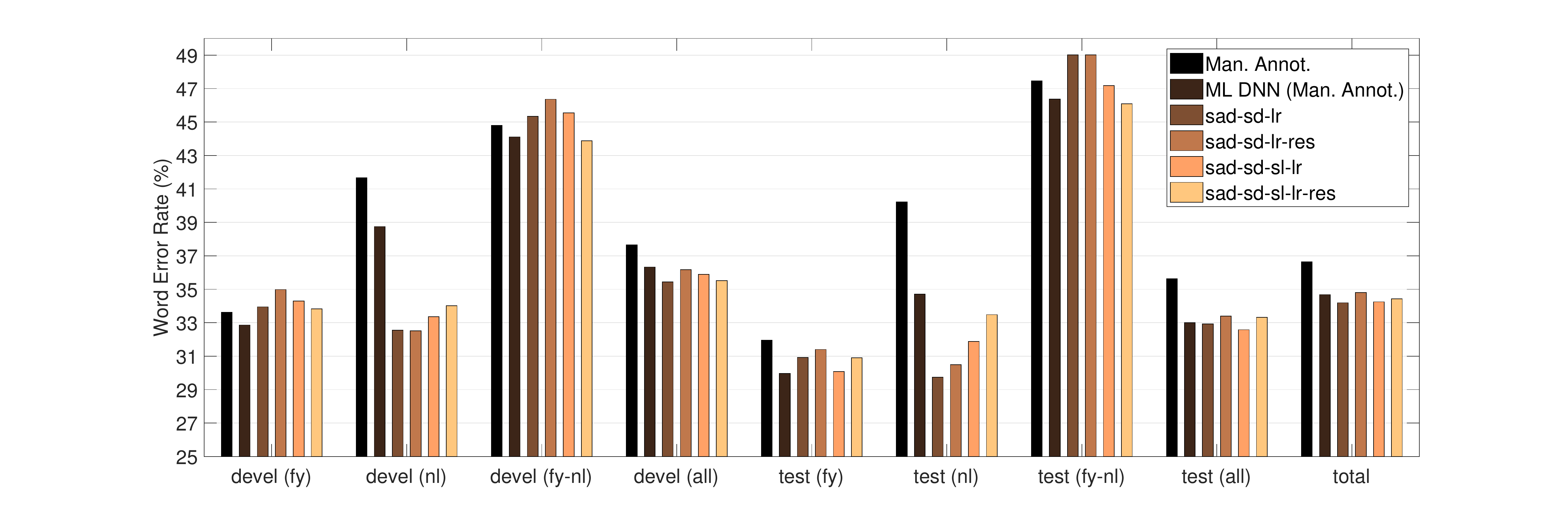}
    }
    \caption{ASR results in WER(\%) for Frisian only (fy), Dutch only (nl), mixed segments (fy-nl) and combined (all) results for the automatic annotation schemes in which speaker labeling is performed after the speech activity detection}
    \label{fig:sd_res}
%    \vspace{-0.35cm}
\end{figure*}

The evaluation is performed on the development and test data of the FAME! speech corpus and the recognition results are reported separately for Frisian only (fy), Dutch only (nl) and mixed (fy-nl) segments. The number of Frisian and Dutch words in the development and test sets are given in Table~\ref{tab:num}. The overall performance (all) is also provided as a performance indicator. The recognition performance of the ASR system is quantified using the word error rate (WER). The word language tags are removed while evaluating the ASR performance.

After the ASR experiments, we compare the CS detection performance of these recognizers. For this purpose, we use a different LM strategy. We train separate monolingual LMs, and interpolated between them with varying weights, effectively varying the prior for the detected language. For each LM, we generate the ASR output for each utterance. Then, we extract word-level segmentation files for each LM weight. By comparing these alignments with the ground truth word-level alignments (obtained by applying forced alignment using the baseline recognizer), a time-based CS detection accuracy metric is calculated. Specifically, we label each frame with a language tag for the ground truth and hypothesized alignments and calculate the total duration of the frames in the reference alignments with a mismatch with the hypothesized language tag. The missed Frisian (Dutch) time is calculated as the ratio of total duration of frames with a Frisian (Dutch) tag in the reference alignment that is aligned to frames without a Frisian (Dutch) tag to the total number of frames with a Frisian (Dutch) tag in the reference alignment. 
\begin{figure*}[ht]
\vspace{-0.35cm}
    \includegraphics[trim=4.5cm 0.5cm 0cm 0cm, width=7in]{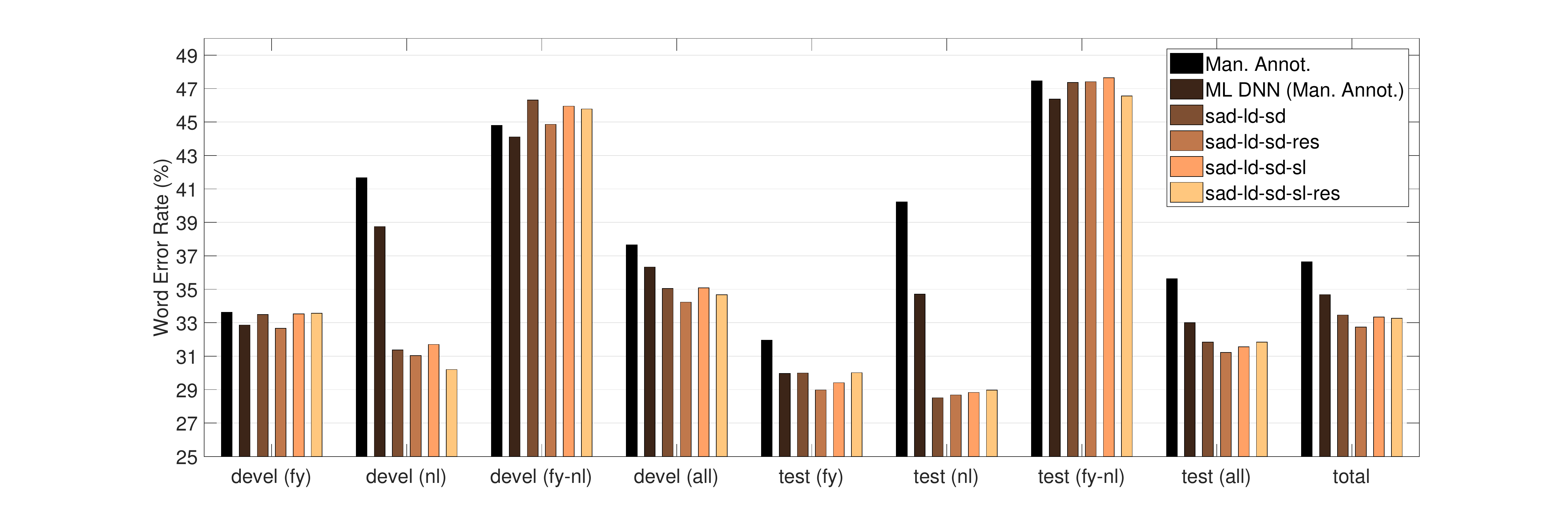}
    \caption{ASR results in WER(\%) for Frisian only (fy), Dutch only (nl), mixed segments (fy-nl) and combined (all) results for the automatic annotation schemes where language labeling is performed after the speech activity detection}
    \label{fig:ld_res}
    \vspace{-0.35cm}
\end{figure*}
\begin{figure*}
    \includegraphics[trim=4.5cm 0.5cm 0cm 0cm, width=7in]{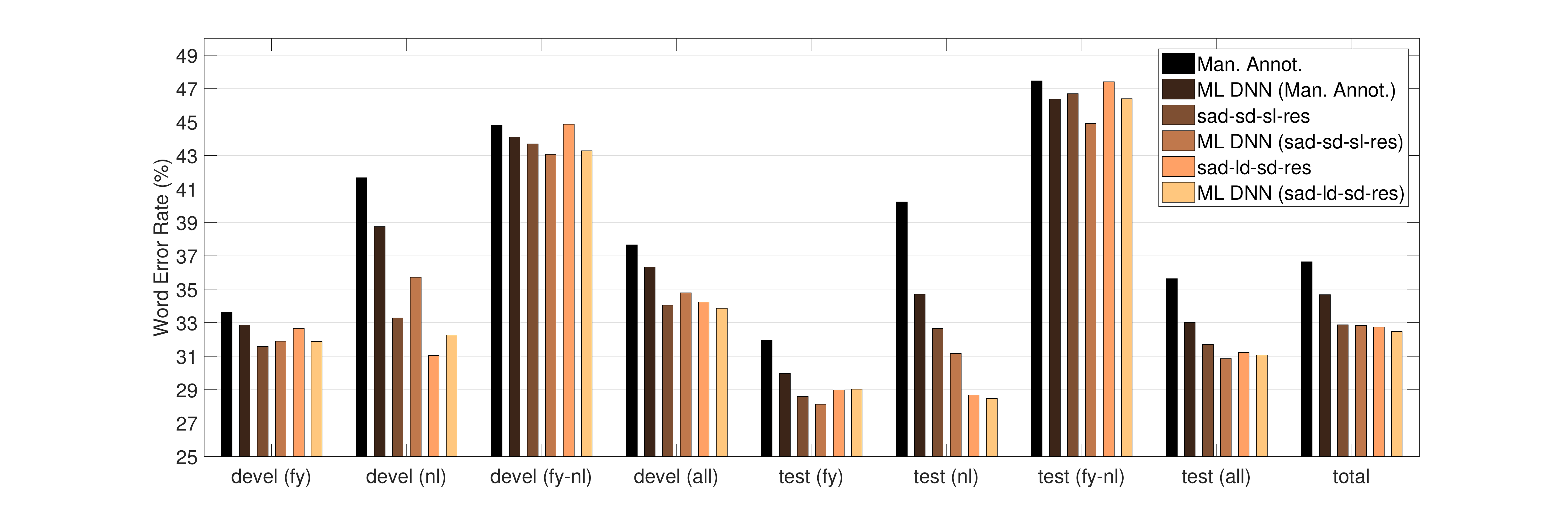}
    \caption{ASR Results in WER(\%) for Frisian only (fy), Dutch only (nl), mixed segments (fy-nl) and combined (all) results - Comparison of single-stage and ML DNN training scheme using the best-performing monolingual and bilingual strategies, namely sad-sd-sl-res and sad-ld-sd-res.}
    \label{fig:fin_res}
    \vspace{-0.35cm}
\end{figure*}

CS detection accuracy is evaluated by reporting the EER calculated based on the detection error tradeoff (DET) graph~\cite{martin1997} plotted for visualizing CS detection performance. The presented code-switching detection results indicate how well the recognizer detects the switches and hypothesizes words in the switched language.

\section{Results}
\label{sec:res}

In this section, we present the results of the ASR and CS detection experiments performed on the development and test data. We first report the detailed WERs for the final bilingual acoustic models trained on the combined data applying standard (single-stage) DNN training. The results in this part are presented in three categories based on the strategy followed for automatic annotation: (1) approaches not using language labeling (sad-sd, sad-sd-res, sad-sd-sl, sad-sd-sl-res), (2) approaches using language recognition (sad-sd-lr, sad-sd-lr-res, sad-sd-sl-lr, sad-sd-sl-lr-res) and (3) approaches using language diarization (sad-ld-sd, sad-ld-sd-res, sad-ld-sd-sl, sad-ld-sd-sl-res). Then, the best-performing monolingual and bilingual strategies are used to train DNN models with ML DNN training by using the combined data in the second stage of the training to compare the impact of single- and two-stage training schemes. Finally, the CS detection performance of these systems are presented to obtain insight about the impact of the automatic annotation and ML DNN training on CS detection accuracy.

\subsection{ASR Results}
\label{ssec:asr}

The WERs obtained on each component of the FAME! development and test data are presented in Figure~\ref{fig:sd_res}. We discuss only the total WER which indicates the ASR performance on both development and test datasets. The first baseline ASR system trained on the manually transcribed data provides a total WER of 36.7\%, while the ML DNN training reduces the total WER to 34.7\%. Adding only automatically transcribed data with pseudo speaker labels (sad-sd) reduces the total WER to 33.6\% providing approximately 1.1\% absolute reduction in the total WER. 

Bilingual rescoring (sad-sd-res) brings minor improvements reducing the total WER of 33.1\%. The best ASR performance is obtained by using the sad-sd-sl-res strategy which also performs speaker linking after the speaker diarization. This approach is henceforth referred to as the best bilingual strategy with a total WER of 32.9\%. The largest performance gains are obtained on the monolingual segments, namely the (fy) and (nl) components of both the development and test data. The results obtained on the mixed segments (fy-nl) demonstrate that these utterances are still very challenging to recognize.

\begin{figure}[t]
\begin{center}
\includegraphics[trim=1cm 1cm 0cm 2cm, width=3.3in]{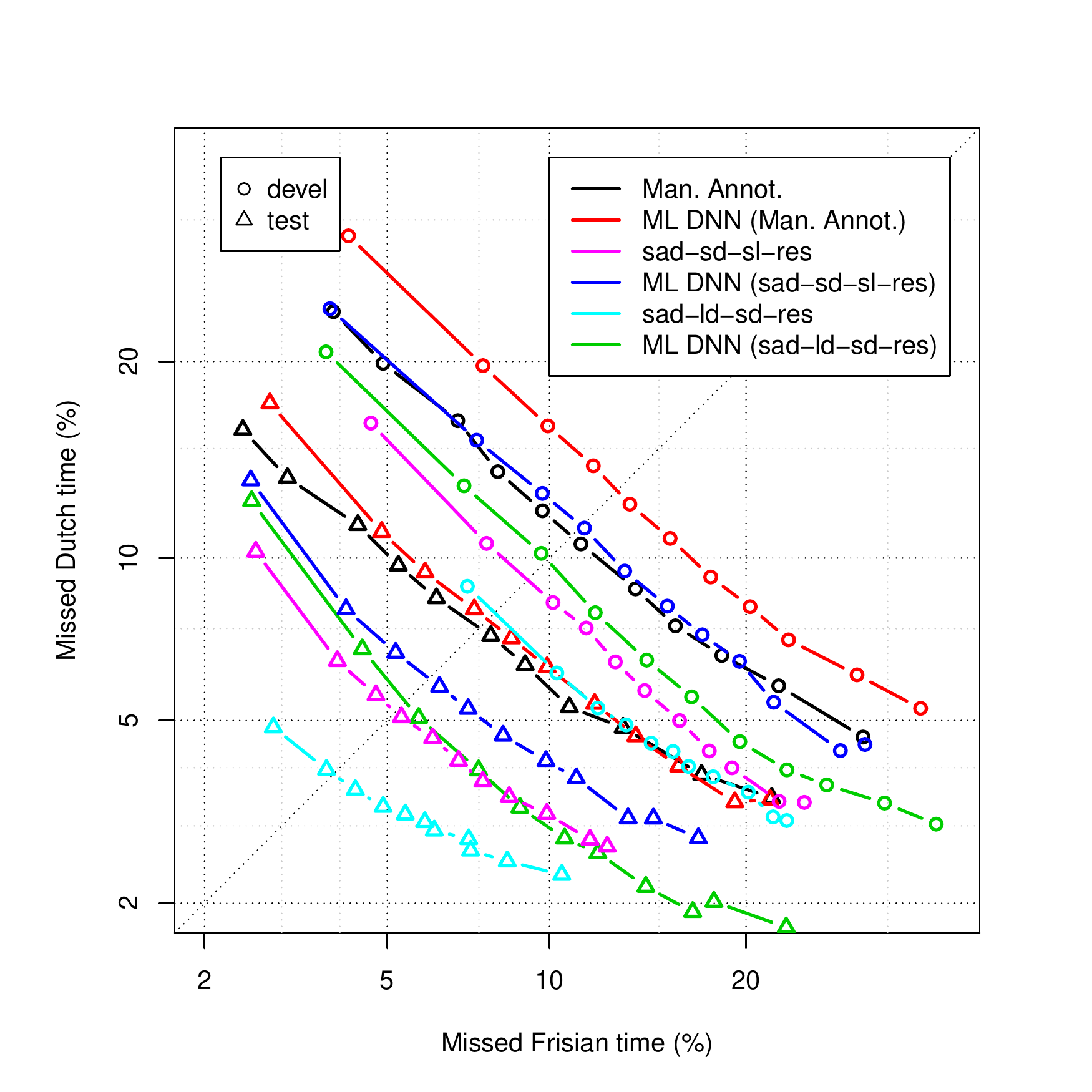}
\end{center}
  \caption{Code-switching detection performance obtained on the FAME! development and test sets}
\vspace{-0.25cm}
\label{fig:det4}
\end{figure}

The automatic annotation strategies applying language recognition after speaker labeling perform worse compared to the ones without language recognition. The annotation accuracy in this setting highly depends on the language homogeneity in speaker-labeled segments, as all segments assigned to a single speaker are tagged with a single language tag. As a result, these strategies are prone to the errors in the language tags. Eventually, these segments are transcribed by the incorrect monolingual ASR resulting in erroneous automatic transcription. Moreover, this monolingual speaker assumption is not very realistic in a CS scenario given that the reference data includes 10 bilingual speakers (all program presenters or reporters).
 
The performance of the ASR systems trained on the automatic annotations generated by first performing language labeling is presented in Figure~\ref{fig:ld_res}. The sad-ld-sd strategy reduces the total WER to 33.4\%, while sad-ld-sd-res with monolingual rescoring further reduces the total WER to 32.7\%.
This is the lowest total WER obtained so far using the combined data for training purposed and it is henceforth referred to as the best-performing monolingual strategy. Speaker linking did not bring further improvements in this scenario yielding 0.5\%-0.6\% higher total WERs than the best-performing monolingual strategy.

Finally, we apply ML DNN training to the best-performing monolingual (sad-ld-sd-res) and bilingual (sad-sd-sl-res) strategies to investigate the effects of using large amounts of multilingual data for DNN training on WER and CS detection. The ASR results with and without ML DNN training are given in Figure~\ref{fig:fin_res}. The ML DNN training yields superior performance compared to the standard DNN training when only the manually annotated data is used for retraining. Retraining with the combined data provides marginal improvements compared to the systems trained only on the combined data. The lowest total WER of 32.5\% is provided by ML DNN (sad-ld-sd-res) strategy.

\subsection{CS Detection Results}
\label{ssec:cs}

The CS detection performance obtained on each component of the FAME! development and test data is presented in Figure~\ref{fig:det4}. The baseline ASR system provides an EER of 10.9\% on the development data and 7.5\% on the test data. Increasing the amount of training data by applying the best-performing bilingual strategy helps with CS detection accuracy reducing the EER to 9.2\% and 5.2\% on the development and test data respectively. Similar to the ASR results, the best-performing monolingual strategy provides the lowest EERs on both sets with 8.1\% on the development and 3.9\% on the test data.

The ML DNN training has an adverse effect on the CS detection accuracy increasing the EERs consistently for each system. The absolute increase in EERs due to the ML DNN training are in the range of 1.8\%-2.1\% on the development data and 0.3\%-0.8\% on the test data.

\begin{table}
\centering
%\vspace{-0.25cm}
\caption{Summary of the main experimental findings}
\vspace{0.25cm}
\addtolength{\tabcolsep}{-5.5pt}
\begin{tabular}{| l | c | c | c | c | c | c |}
\hline
\parbox[t]{2cm}{ASR System} & \multicolumn{3}{c|}{\parbox[t]{2.4cm}{\centering ASR results in WER (\%)}} & \multicolumn{3}{c|}{\parbox[t]{3cm}{\centering CS detection results in EER (\%)}} \\
\hline
  & \parbox[t]{0.8cm}{\centering Dev.} &  \parbox[t]{0.8cm}{\centering Test}  & \parbox[t]{0.8cm}{\centering Tot.} & \parbox[t]{1cm}{\centering Dev.} & \parbox[t]{1cm}{\centering Test}  & \parbox[t]{1cm}{\centering Tot.}\\
\hline \hline
\parbox[t]{2cm}{Man. Annot} & 37.7 & 35.6 & 36.7 & 10.9 & 7.5 & 9.2\\
\hline
\parbox[t]{2cm}{+ML DNN} & 36.3 & 33.0 & 34.7 & 12.9 & 7.8 & 10.3\\
\hline \hline
\parbox[t]{2cm}{sad-sd-sl-res} & 34.1 & 31.7 & 32.9 & 9.2 & 5.2 & 7.1\\
\hline
\parbox[t]{2cm}{+ML DNN} & 34.8 & 30.8 & 32.8 & 11.3 & 6.0 & 8.6\\
\hline \hline
\parbox[t]{2cm}{sad-ld-sd-res} & 34.2 & 31.2 & 32.7 & 8.1 & 3.9 & \bf{5.9}\\
\hline
\parbox[t]{2cm}{+ML DNN} & 33.9 & 31.1 & \bf{32.5} & 9.9 & 5.5 & 7.6\\
\hline
\end{tabular}
%\vspace{-0.25cm}
\label{tab:summ}
\end{table}

\section{General Discussion}
\label{sec:dis}
A summary of the ASR and CS performance of the described approaches is given in Table \ref{tab:summ}. In the following sections, we provide a detailed discussion of the results presented in the previous section and list possible future directions. 

\subsection{ASR Performance}
\label{ssec:dis_asr}
With the motivation of training better bilingual acoustic models for CS speech in a semi-supervised manner, various automatic annotation strategies have been investigated on Frisian-Dutch broadcast data. Motivated by the effectiveness of speaker-adaptive training in the target application, automatic speaker labeling has also been used in these approaches by performing speaker diarization and in some cases speaker linking to actually identify frequently appearing speakers. Due to the bilingual nature of the data, a subset of the described strategies has relied on language recognition in the form of either language identification or diarization. The language-specific approaches have incorporated monolingual ASR resources at the back-end unlike the bilingual ASR resources used by the strategies without language recognition.

The ASR results indicates the advantage of the proposed semi-supervised acoustic model training schemes for improving bilingual ASR performance by automatically annotating a relatively small portion of the Frisian-Dutch broadcast archive (less than 10\%) and training acoustic models on the combined (reference and automatically annotated) data. The best-performing bilingual strategy (sad-sd-sl-res) results in an absolute reduction of 3.8\% and 1.8\% in the total WER compared to the baseline systems trained only on the manually annotated data applying either a single-stage (standard) or two-stage (ML) training, respectively. These gains can be increased respectively up to 4\% and 2\% absolute by employing language diarization as done in the sad-ld-sd-res strategy.

While some strategies have been found to be effective, some others have provided marginal or no improvements, namely the strategies where the language tagging is preceded by speaker tagging. From these results, it can be concluded that the assumption of language-homogeneity in the segments tagged with the single speaker label does not hold in CS scenarios. This is conceivable given that the reference data includes multiple bilingual speakers. Moreover, conditioning language tags on the speaker tags makes the annotation strategy more vulnerable to the possible errors in the speaker diarization system, resulting in a larger number of segments transcribed by the mismatched monolingual ASR resources.

Applying ML DNN training on the combined data has also been explored with the aim of obtaining more reliable shared hidden layers on a large amount of manually annotated multilingual data. In this case, the impact of ML DNN training on the ASR performance was limited. This is due to the increased amount of training data used in the retraining stage. Therefore, the performance gains are marginal compared to the gains reported only using the reference data.

The training data created using the proposed techniques enables the use of more data from the high-resourced mixed language (Dutch in this case) during the training of the bilingual AM. In this way, further improvements in bilingual ASR performance have been reported in more recent experiments \cite{yilmaz2018} by incorporating recurrent architectures trained on the created CS training data and more Dutch data from other databases.
 
\subsection{CS Detection Accuracy}
\label{ssec:dis_csdet}

CS detection is an interesting application that reveals the language boundaries in the CS speech. The influence of using the automatically annotated data and ML DNN training has been explored in the last stage of the experiments. One clear outcome is the superior CS detection performance of the best-performing monolingual strategy over the other systems. The automatic annotation scheme using monolingual resources at the back-end (sad-ld-sd-res) brings a considerable EER reduction indicating the higher quality of the language tags assigned by the final bilingual ASR system trained on the combined data.

The detrimental effect of the ML DNN training on the CS detection is consistently observed in all scenarios. It is worth mentioning that CS detection accuracy also reduces after the ML DNN training using only the manually annotated data contrary to the gains obtained in ASR performance reported earlier. We can conclude from these results that including an unseen language in the DNN training procedure reduces the quality of assigned language tags by a bilingual ASR system. Therefore, refraining from applying ML DNN training is good practice when the bilingual acoustic models are (also) used for the detection of language switches.

\subsection{Future Directions}
\label{ssec:dis_fut}

The experimental results reported in this work show the potential of using semi-supervised acoustic model training in bilingual scenarios, most notably for the ASR of CS speech, in the presence of speaker and language recognition systems. One straightforward extension to the presented work is investigating the impact of scale-ups in the amount of raw broadcast data on the performance gain. In our project, this investigation is postponed until the rest of the broadcast archive is digitalized.

Increasing the amount of training CS speech, the proposed bilingual acoustic model training scheme will make the use of more training data from the high-resourced mixed language viable. This is expected to bring further improvements in the ASR accuracy of the CS speech. Finally, integrating more dedicated language models that can enable language switches in a more flexible manner will bring complementary information to the acoustic evidence from the acoustic models developed in this work.

Data selection techniques have been often presented as a part of previously presented semi-supervised training techniques. These techniques use a criterion that serves as a confidence measure for the hypothesized transcription to eliminate the segments with \textit{unreliable} automatic annotations. Another future step is to introduce such confidence measures in the proposed framework to explore the quality of the final acoustic models.

\section{Conclusions}
\label{sec:conc}
This paper describes several automatic annotation strategies using language and speaker recognition for semi-supervised bilingual acoustic model training. Specifically, raw speech segments extracted using a speech activity detection system are automatically labeled with speaker and language information and later used for acoustic model training together with the reference data with manual annotations. We compare the ASR and CS detection performance of these acoustic models with the baseline systems trained only on the manually annotated data. Applying language diarization to the speech-only segments to assign language labels followed by speaker labeling using a speaker diarization system has been found to be the best automatic annotation scheme for delivering the bilingual acoustic model with the highest ASR and CS detection performance. 

Multilingual DNN training has been previously shown to significantly improve the accuracy of the bilingual ASR system, partly compensating for the limited resources of the Frisian language. The CS detection results presented in this work indicate the adverse effect of the multilingual DNN training on the quality of the language tags assigned by the bilingual ASR system. In addition, the two-stage multilingual training is found to be ineffective in improving the ASR performance after using the automatically annotated data for training purposes, and it can be omitted due to the increased amount of Frisian-Dutch training data.

\section{Acknowledgements}
\label{sec:ack}
This research has been funded by the NWO Project 314-99-119 (Frisian Audio Mining Enterprise) and SRI International. We would like to thank Aaron Lawson for his support with the arrangement of this research visit; Martin Graciarena for his contributions to the speech activity detection system; Diego Castan for his contribution to the language diarization system; Mahesh Nandwana for his contribution to the speaker recognition system; Jelske Dijkstra and Hans Van de Velde for their contribution to the section on the Frisian language; and Frederik Kampstra for delivering the data from the Omrop Frysl\^{a}n archive.

\bibliography{refs}

\end{document}